\documentclass{article}

\usepackage{arxiv}

\usepackage[utf8]{inputenc} 
\usepackage[T1]{fontenc}    
\usepackage{hyperref}       
\usepackage{url}            
\usepackage{booktabs}       
\usepackage{amsfonts}       
\usepackage{nicefrac}       
\usepackage{microtype}      
\usepackage{lipsum}		
\usepackage{graphicx}
\usepackage[numbers]{natbib}
\usepackage{doi}
\usepackage{amsmath}
\usepackage{amssymb}

\usepackage{subfig}
\usepackage{enumitem}
\usepackage{tabularx}
\usepackage{csquotes}
\usepackage{url}
\usepackage{xspace}
\usepackage{booktabs}
\usepackage{lineno}
\usepackage{algorithm} 
\usepackage{algorithmicx}%
\usepackage[noend]{algpseudocode}%

\newcommand{\ModelBaseline}{\mathcal{B}}
\newcommand{\ModelReduced}{\mathcal{R}}
\newcommand{\ModelSemRoom}{\mathcal{SR}}
\newcommand{\ModelSemElem}{\mathcal{SE}}
\newcommand{\ModelSemRoomElem}{\mathcal{SRE}}
\newcommand{\fn}[2]{\textsc{#1}\!\left(#2\right)}
\newcommand{\I}[1]{\text{#1}}

\begin{document}

\title{Automating Computational Design with Generative AI}

\author{ \href{https://orcid.org/0000-0002-6320-8891}{\includegraphics[scale=0.06]{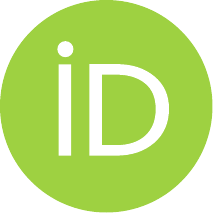}\hspace{1mm}Joern Ploennigs}\\
	University of Rostock, AI for Sustainable Construction\\
	Justus-v.-Liebig-Weg 6\\
	Rostock, 18059, Germany \\
	\texttt{Joern.Ploennigs@uni-rostock.de} \\
	\And
	\href{https://orcid.org/0000-0001-5232-0039}{\includegraphics[scale=0.06]{orcid.pdf}\hspace{1mm}Markus Berger} \\
	University of Rostock, AI for Sustainable Construction\\
	Justus-v.-Liebig-Weg 6\\
	Rostock, 18059, Germany \\
	\texttt{Markus.Berger@uni-rostock.de} \\
}

\date{May 2024}
\maketitle

\begin{abstract}
AI image generators based on diffusion models have recently garnered attention for their capability to create images from simple text prompts. However, for practical use in civil engineering they need to be able to create specific construction plans for given constraints. This paper investigates the potential of current AI generators in addressing such challenges, specifically for the creation of simple floor plans. We explain how the underlying diffusion-models work and propose novel refinement approaches to improve semantic encoding and generation quality. In several experiments we show that we can improve validity of generated floor plans from 6\,\% to 90\,\%. Based on these results we derive future research challenges considering building information modelling. With this we provide: (i) evaluation of current generative AIs; (ii) propose improved refinement approaches; (iii) evaluate them on various examples; (iv) derive future directions for diffusion models in civil engineering. 
\end{abstract}

\keywords{
generative AI \and
diffusion models\and
deep learning\and
computational design\and
building information modelling
}



\setlist{nolistsep}

\section{Introduction}

The recent releases of AI-based art generators like DALL$\cdot$E 2, Midjourney and Stable Diffusion and text generators like ChatGPT and GPT-4 are sparking discussions about how they can help architects and planners in their daily business. These tools are already finding widespread adaptation with for example 7\,\% of all Midjourney queries being related to architecture \cite{ploennigs2023ai}. Of those prompts 43\,\% are related to floor plans. These generative AI tools show a remarkable capability in generating architectural images with a higher degree of creativity and good understanding of architectural styles, as shown in \cite{ploennigs2023generative}. This raises the question of whether they can be used for more challenging tasks, like generating floor plans or sectional drawings, which are nowadays often manually drawn on paper or in CAD (Computer Aided Design) tools. 

The process of planning for construction projects in civil engineering is still very manual and time-consuming, with architects and engineers spending numerous hours creating designs and plans \cite{van2023impact}. There is enormous potential for boosting productivity by 50\,\% to 60\,\%, which could result in an extra 1.6 trillion USD being added annually to the value of the construction industry \cite{barbosa2017reinventing}. AI approaches are considered a core enabler of this productivity boost \cite{PAN2021103517,pan2023integrating,debrah2022artificial}. 

The recent breakthroughs of generative AI tools for images are based on the introduction of \textit{diffusion models} in deep learning. They were first introduced by Sohl-Dickstein et al. in \cite{sohl2015deep} and are nowadays the main approach for image and video generation (see Section~\ref{sec:diffusionmodels}). In this paper, we want to investigate how these \textit{diffusion models} can be used for computational design of architectural drawings like floors plans and how we can improve their performance for this task. To achieve this, we do the following: 
\begin{itemize}
    \item Evaluate the capability of diffusion models in computational design
    \item Discuss the requirements of diffusion models for architectural planning 
    \item Propose the fine-tuning of diffusion models with domain examples 
    \item Propose a novel semantic encoding of floor plans for improved fine-tuning
    \item Fine-tune a model for each proposed encoding 
    \item Compare the performance and workflows for those encodings 
    \item Discuss if and how we can directly use BIM data for diffusion models  
    \item Derive the requirements and sketch the architecture for a BIM-based diffusion model 
    \item Identify future directions for diffusion models in computational design 
\end{itemize}

We use the openly available, very commonly used Stable Diffusion v2.1 model as our basis for fine-tuning, to be able to realistically evaluate the capabilities of current diffusion models. To improve its results, we then propose a novel approach to fine-tune the diffusion models with floor plans rendered in specific visual styles to enable semantic encodings. Based on our experiments, we also identify the limitations of current approaches and sketch out the option of a BIM-based diffusion model as a potential solution, without implementing it, as this is beyond the scope of the paper and thus future work.

\section{State of the Art in Computational Design}\label{sec:soa}
The idea of automating design steps with technology is an old one. Various approaches developed over time, from \textit{Parametric} to \textit{Generative}, over \textit{Algorithmic Design}. Caetano et al. grouped them the under the term \textit{Computational Design} \cite{caetano2020computational}.

The idea of \textit{Parametric Design} is that the parameters defining the shape (e.\,g.\ length, height, ...) of components are varied by functions in a computer, in contrast to being set directly \cite{gursel2012creative,jabi2013parametric,schnabel2007parametric}. It was already used before computers were introduced e.\,g.\ by Antoni Gaudi for the Sagrada Familia \cite{burry1993expiatory}. Parametric design is used to generate variants of structural elements that are then validated and ranked structurally \cite{monedero2000parametric,rempling2019automatic} or in their energy performance\cite{granadeiro2013building}.

\textit{Generative Design} is using evolutionary or agent-based algorithms to generate whole designs \cite{caldas2008generation,krish2011practical}. 
The challenge in those approaches is that they are bound in their generation capability by the underlying generation rules and representations (e.\,g.\ genome encoding) and cannot recombine them in creative ways. Further they are prone to generate random results outside of the requested targets due to their
probabilistic or non-deterministic nature of their search \cite{caetano2020computational,chaszar2016generating}.

\textit{Algorithmic design} approaches usually use rules or code grammars to create designs \cite{oxman2017thinking,caetano2019integration,sun2022wallplan}. Recent approaches combine such a shape grammar with reinforcement learning to generate energy efficient designs \cite{mandow2020architectural}. These approaches are deterministic, but, require understanding of the rules/code to control the generation process, which makes them less applicable in practise.

A common use case across all approaches is the generation of floor plans\cite{weber2022automated}. Recently, also Generative Adversarial Networks, a Deep Neural Network architecture, are used to generate floor plans \cite{chang2021building, wu2022generative, nauata2021house}. They are similar to diffusion models, which we will introduce in the next section.

\section{Introduction to Diffusion models }
\subsection{Model Structure }\label{sec:diffusionmodels}

\textit{Diffusion Models} revolutionized the area of generating art with AI. They were first introduced in \cite{sohl2015deep} and are inspired by nonequilibrium thermodynamics.  Diffusion-models are based on a Deep Neural Network (DNN) architecture that does two things: First, it successively destroys the information in the image it is trained on by adding white noise in a so called \textit{forward diffusion} step.  In each step it trains a DNN to be able to recover the lost information in a \textit{reverse diffusion} step. The resulting image now contains some of the original information, like style and colour, and some newly added information which is associated to tokens describing the image. As the DNN is only trained to repair a small piece of lost information in the image, the generation approach becomes more stable and thus solves a major challenge with traditional approaches, which try to generate the whole image from scratch and often fail in attaining decent results. 

Formally, they are formulated using a Markov chain of $T$ steps, meaning that each step only depends on the previous one. Given a pixel $x_0$ of an original image with distribution $q(x_0)$, we define a forward diffusion process by adding Gausian noise with variance $\beta_t$ to $x_{t-1}$ with mean $\mu_t$ derived from $q(x_{t-1})$. This produces a new latent variable $x_t$ with distribution
\begin{equation}
    q(x_{t} | x_{t-1}) = N(x_t; \mu_t=\sqrt{1-\beta_t}x_{t-1}; \beta_t).
\end{equation}
The trick now is, that if the steps, i.\,e.\ the variance $\beta_t$, are small enough, then the reverse diffusion can be assumed to be a Gaussian process for which the variance $\hat\beta_\theta$ depends only on $\beta_t$ and on a mean function $\hat\mu$ as follows
\begin{equation}
    p(x_{t-1} | x_{t}) = N(x_{t-1}; \hat\mu(x_t, x_0 | P); \hat\beta_\theta).
\end{equation}
This mean function $\hat\mu(x_t, x_0 | P)$ is now learned by a Deep Neural Network (DNN), into which we also feed the vector encoding $P$ of the tokens in a prompt describing the image\cite{ho2020denoising}. 

\begin{figure}[htb]%
    \centering
    \subfloat[\centering Step 1]{{\includegraphics[width=0.32\columnwidth]{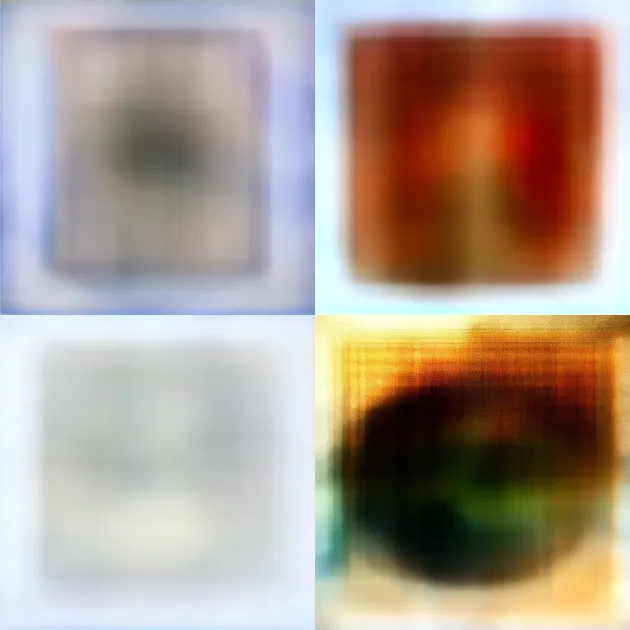} }}%
    \subfloat[\centering Step 2]{{\includegraphics[width=0.32\columnwidth]{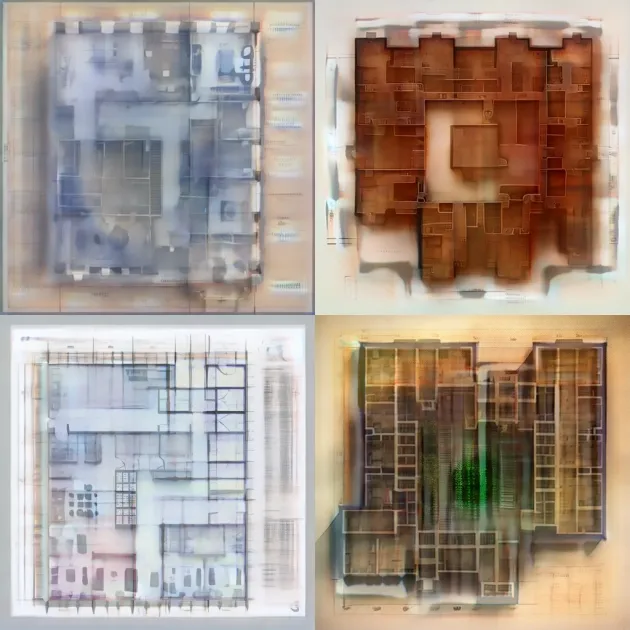} }}%
    \subfloat[\centering Step 3]{{\includegraphics[width=0.32\columnwidth]{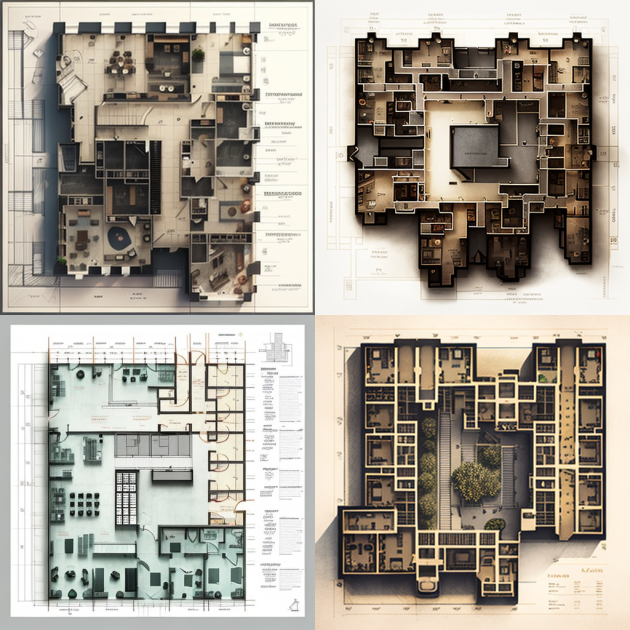} }}%
    \caption{Examples of iterative denoising steps for a building floor plan in Midjourney v4.}%
    \label{fig:gen_levels}%
\end{figure}


The recent generation of Diffusion Models is using the CLIP (Contrastive Language Image Pre-training) architecture presented in \cite{radford2021learning} and \cite{ramesh2021zero}. It is not simply conditioning the mean function $\hat\mu(x_t, x_0 | P)$ on the tokens vector $P$ describing the image, but learns how tokens are linked to image parts in the diffusion model. This additional abstraction layer allows CLIP to learn object types and image composition on its higher layer, and details such as surface materials and colours on the deeper diffusion layer. An example of this step-by-step process is shown in Fig.~\ref{fig:gen_levels}.


\subsection{Fine-tuning Diffusion Models}\label{sec:tuning}
A consequence of training the diffusion model on the tokens vector $P$ is that the model can only generate images using these known tokens. This is not necessarily an issue as large diffusion models like DALL$\cdot$E~2, Midjourney and Stable Diffusion are trained on a large variety of tokens. But specific applications like designing floor plans are not well represented, leading to bad performance as shown in section\ \ref{sec:experiments}. 
Fortunately, this new generation of diffusion models allows us to refine their federated models with a few example images from a desired application domain, which is called fine-tuning or few-shot-learning. 
There are multiple possible approaches for fine-tuning. The first is to retrain the whole diffusion-model on the new dataset and thus refine the weights in the deep neural network (DNN) \cite{Ruiz_2023_CVPR}. This creates a variant of the initial federated diffusion-model that is of the same size as the original network. However, it is quite computationally expensive to refine all the weights of such a DNN. To avoid this, other approaches introduce new layers into the federated diffusion model and modify only their weights. This results in smaller and faster to train models, as the original federated model is not modified and fewer weights need to be fine-tuned \cite{hu2021lora, roich2022pivotal}. 
The final approach is to not modify the DNN, but the input token vector $P$ from the prompt. The text prompt sent to the diffusion model needs to be encoded as a number vector in order to be used as an input layer for the DNN. This fine-tuning approach only changes the weights of this word embedding vector, a process called \textit{textual inversion} \cite{gal2022textual}. The benefit of this approach is that it allows for very fast training and results in a very small fine-tuned model that only contains the modified word embedding vector. We will use textual inversion for fine tuning the diffusion models later in this paper as it allows significant improvements of results with fast retraining speeds and small models.



\section{Methodologies for Generative Design }
\subsection{Idealized Integrated Workflow}

One goal of this paper is to investigate how generative AI tools can support practitioners in their typical design workflows. In this section, we want to discuss an idealized workflow that, in our opinion, fully exploits the potential of generative AI tools in the design process. We will then also discuss different approaches to making aspects of this workflow feasible with currently existing models. For each of these approaches we will then describe the training process for a simple floor plan generation model in section~\ref{sec:finetuning} and quantitatively evaluate these models in section~\ref{sec:experiments}. 

\begin{figure}[htb]%
    \centering
    \includegraphics[width=0.75\columnwidth]{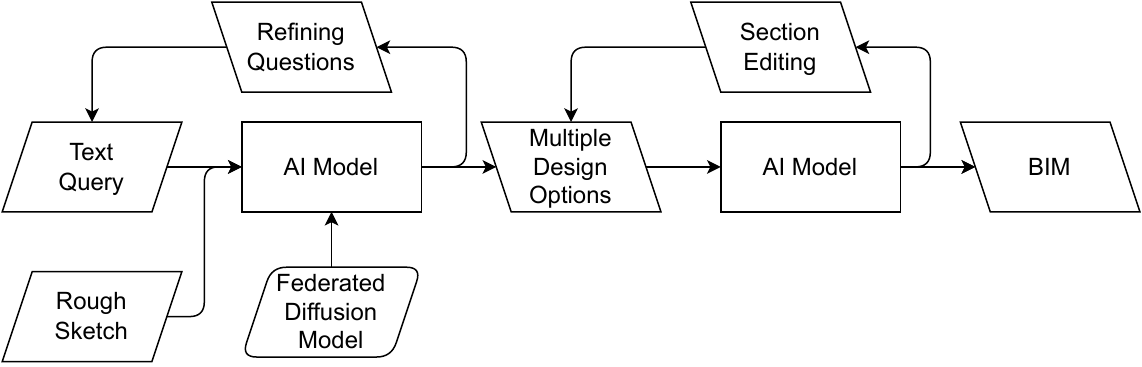} 
    \caption{Idealized integrated workflow for AI-based design}%
    \label{fig:ideal_workflow}%
\end{figure}

In this workflow shown in Figure~\ref{fig:ideal_workflow}, a planner uses a text interface or speech to text interface to specify the keywords describing their targeted design. As an alternative input, the planner can sketch out a rough plan to provide e.\,g.\ the general shape of a more complex planning tasks. The AI system would refine these initial inputs by generating questions to collect additional requirements and clarify uncertainties in a dialog with the user. These requirements may encompass functional (e.\,g. room types, size, etc.), non-functional (e.\,g. materials), stylistic (e.\,g. shape and colours), or computational constraints (e.\,g. structural dynamics, energy-efficiency). It would then generate multiple design options using a federated diffusion model, validate and rank them and allow the user to select the best variants and edit areas that require more work. The final design should be exported to a BIM model like IFC, which users can easily import into their tools for further use. This workflow is based on common workflows for image diffusion models in the AI art domain \cite{ploennigs2023ai}. We consider it a good fit for civil engineering, as it allows a user to to control the design process in multiple ways, from specifying and refining requirements, to editing model parts.


\subsection{Existing Out-of-the-Box Workflow (Model $\ModelBaseline$)}
Examples of existing diffusion models in common use are Midjourney, DALL$\cdot$E by OpenAI, Google Imagen and Stable Diffusion. All models are available as cloud services, which pose a privacy and data security risk for designers, as the prompts need to be sent to external servers. Only Stable Diffusion is an open-source model that can be installed and run locally on a computer. It also can be fine-tuned to adopt it to specific datasets as shown in this paper. 

\begin{figure}[t]%
    \centering
    \includegraphics[width=0.75\columnwidth]{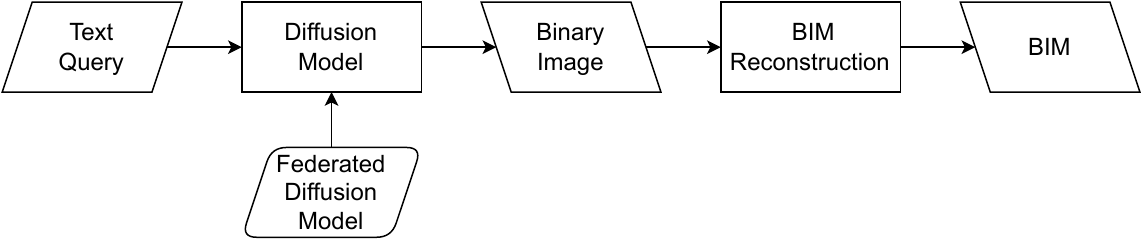}
    \caption{Generation Workflow for out-of-the-box Diffusion Model}%
    \label{fig:general_binary_model}
\end{figure}

The most immediate way a user testing the technology will engage with it, is to run one of the pre-trained, out-of-the box models to generate their floor plans. We call this approach our baseline approach $\ModelBaseline$, which is described in Fig.~\ref{fig:general_binary_model}. The user writes a text query describing their design idea and sends it to the generative AI.The generative AI then uses a federated diffusion model and generates one or multiple binary images as a response to the query. 

This workflow is different to our idealized workflow in Fig.\ \ref{fig:ideal_workflow} in multiple ways. There is no refinement loop that generates clarifying questions or allows us to edit sections, as they are not currently supported by the default models \cite{ploennigs2023ai}. Further, the out-of-the-box approach creates only a binary image of a floor plan instead of directly generating a BIM model. In theory, this binary image could be used to automatically reconstruct a BIM model. However, this is a non-trivial problem with ongoing research \cite{gimenez2016automatic,zeng2019deep,pizarro2022automatic}. In practise, the user still needs to manually create BIM models from the generated images.

\subsection{Proposed Fine-tuning Approach (Model $\ModelReduced$)}

\begin{figure}[t]%
    \centering
    \includegraphics[width=0.3\columnwidth]{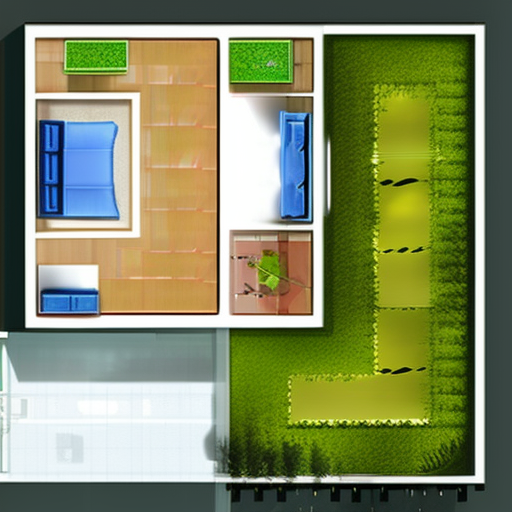}
    \caption{Example floor plan of a house with garden created by the out-of-the-box model $\ModelBaseline$.}%
    \label{fig:example_floor_plan}
\end{figure}

To improve the generated results, we propose to fine-tune the out-of-the-box diffusion model with training images that follow a unified visual style and training prompts that accurately represent semantic properties such as the shown number and type of rooms. The out-of-the-box approach  $\ModelBaseline$ is prone to creating faulty results, as we will confirm in the experimental section\ \ref{sec:experiments}. Fig.~\ref{fig:example_floor_plan} shows an example floor plan generated by Stable Diffusion v2.1 for the query \enquote{building floor plan of a one family house with a garden}. The image shows some resemblance to a floor plan, but also many incorrect elements. The reason for the low-quality result is that the out-of-the-box model is trained on a wide variety of images and will thus contain a large amount of visual variance instead of the expected narrow focus on floor plans.

\begin{figure}[t]%
    \centering
    \subfloat[\centering Training]{{\includegraphics[width=0.75\columnwidth]{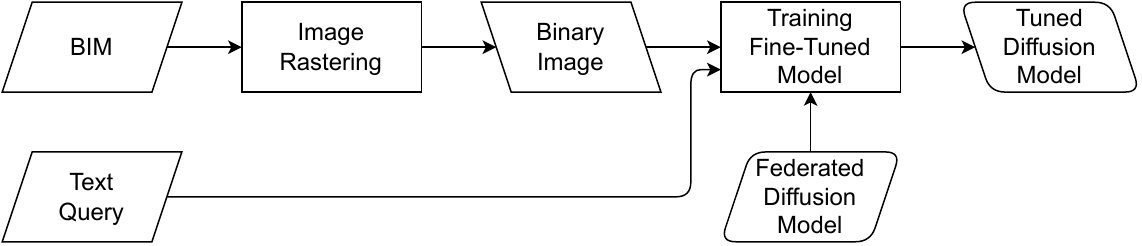} }}\\%
    \subfloat[\centering Generation]{{\includegraphics[width=0.75\columnwidth]{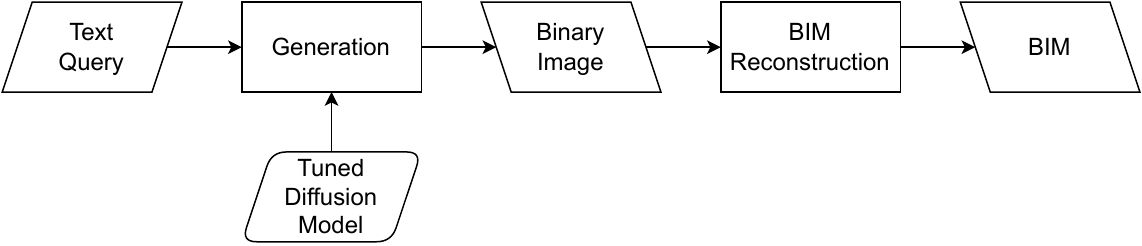} }}%
    \caption{Workflow for a refined 2D/3D bitmap-based Diffusion Model}%
    \label{fig:refined_binary_model}%
\end{figure}

Therefore, it is necessary to fine-tune the diffusion model to train it to understand what specific kinds of floor plans are expected. This can be done by retraining the model with new sample images and queries that resemble the expected results as discussed in section\ \ref{sec:tuning}. For this step, we utilize the first of three strategies we will describe throughout this section:

\begin{description}
\item[Simplify:] The more different semantics a floor plan contains, the harder it is for the diffusion model to differentiate between them. Thus, it is important to remove unnecessary elements (furniture, assets, etc.) and focus on the semantics relevant for floor planning (walls, doors, and windows). The removed unnecessary elements may be added in a following in-painting step in the section editing step from Fig.\ \ref{fig:ideal_workflow}. This strategy will be our baseline for the retraining of model $\ModelBaseline$, which will be called $\ModelReduced$ (Reduced).
\end{description}
This modifies our workflow as illustrated in Fig.~\ref{fig:refined_binary_model}. We now have a training and a generation phase. In the training phase we fine-tune an existing federated diffusion model with a small set of sample images and queries. We then use the tuned diffusion model in the generation phase to generate designs that are closer to the training data. 


\subsection{Proposed Improved Fine-Tuning with Color Encoding of Rooms (Model $\ModelSemRoom$) and Elements (Model $\ModelSemElem$)}

\begin{figure}[t]%
    \centering
    \subfloat[$\ModelReduced$]{{\includegraphics[width=0.24\columnwidth]{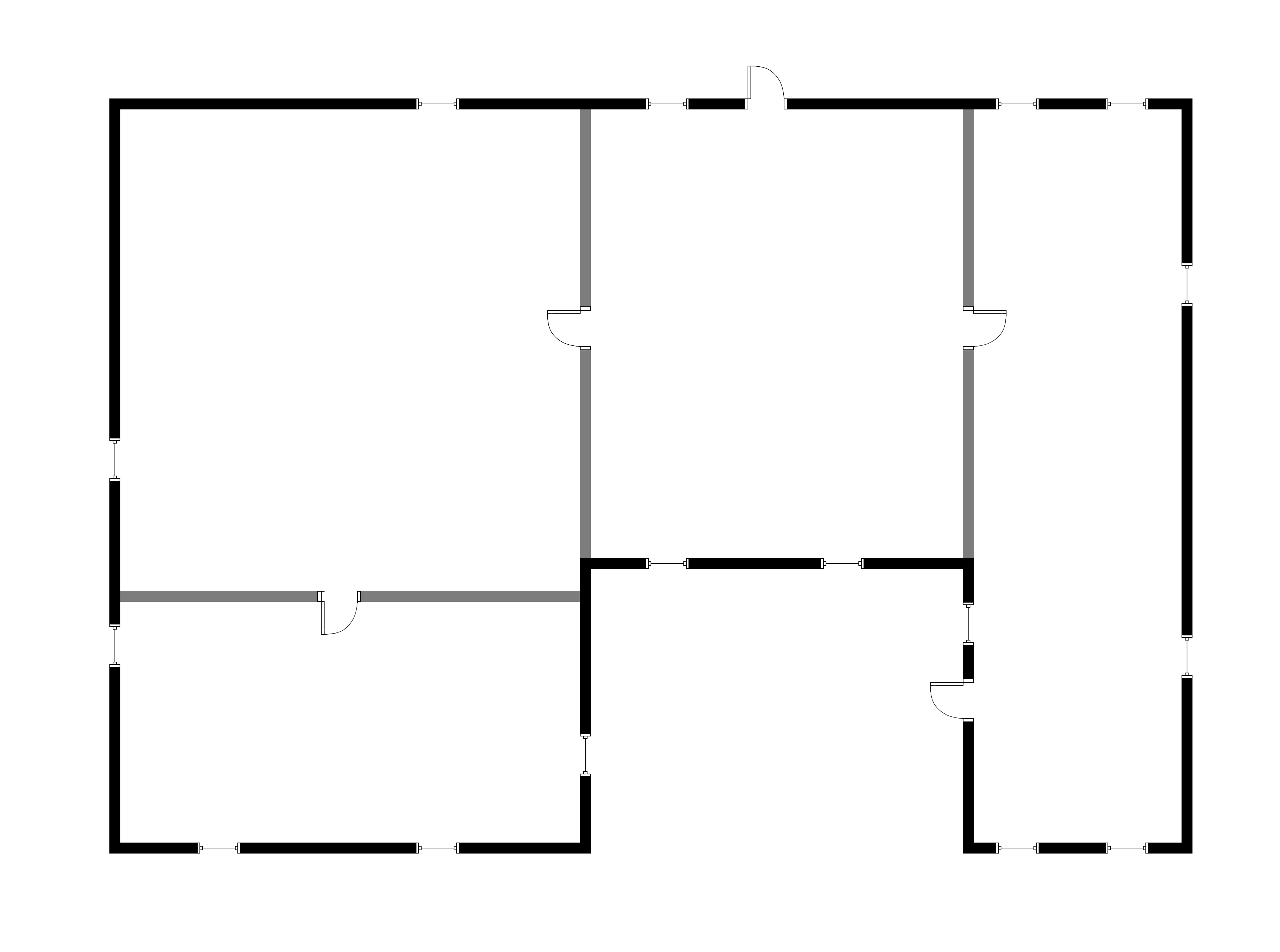} }}%
    \subfloat[$\ModelSemRoom$]{{\includegraphics[width=0.24\columnwidth]{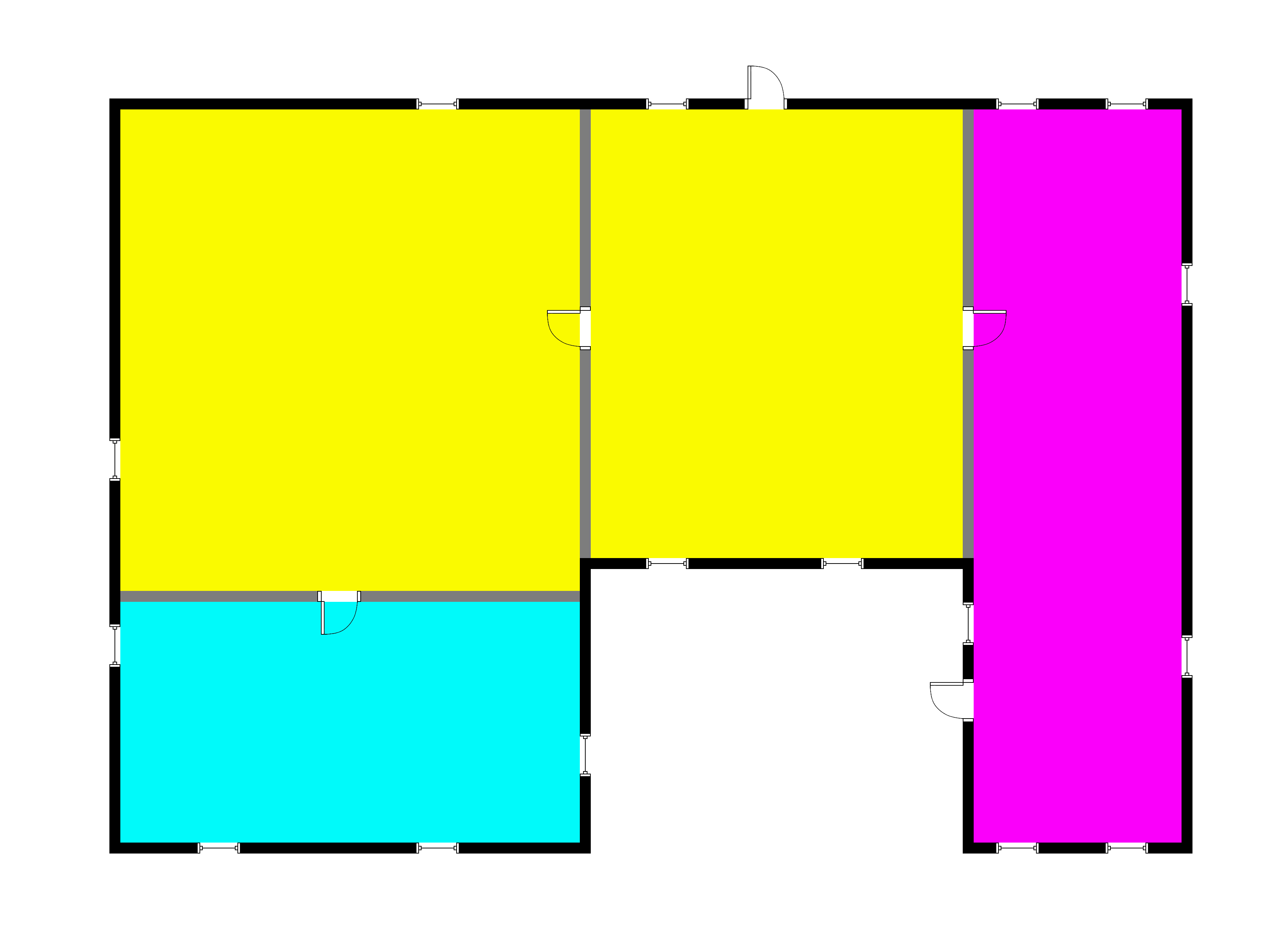} }}
    \subfloat[$\ModelSemElem$]{{\includegraphics[width=0.24\columnwidth]{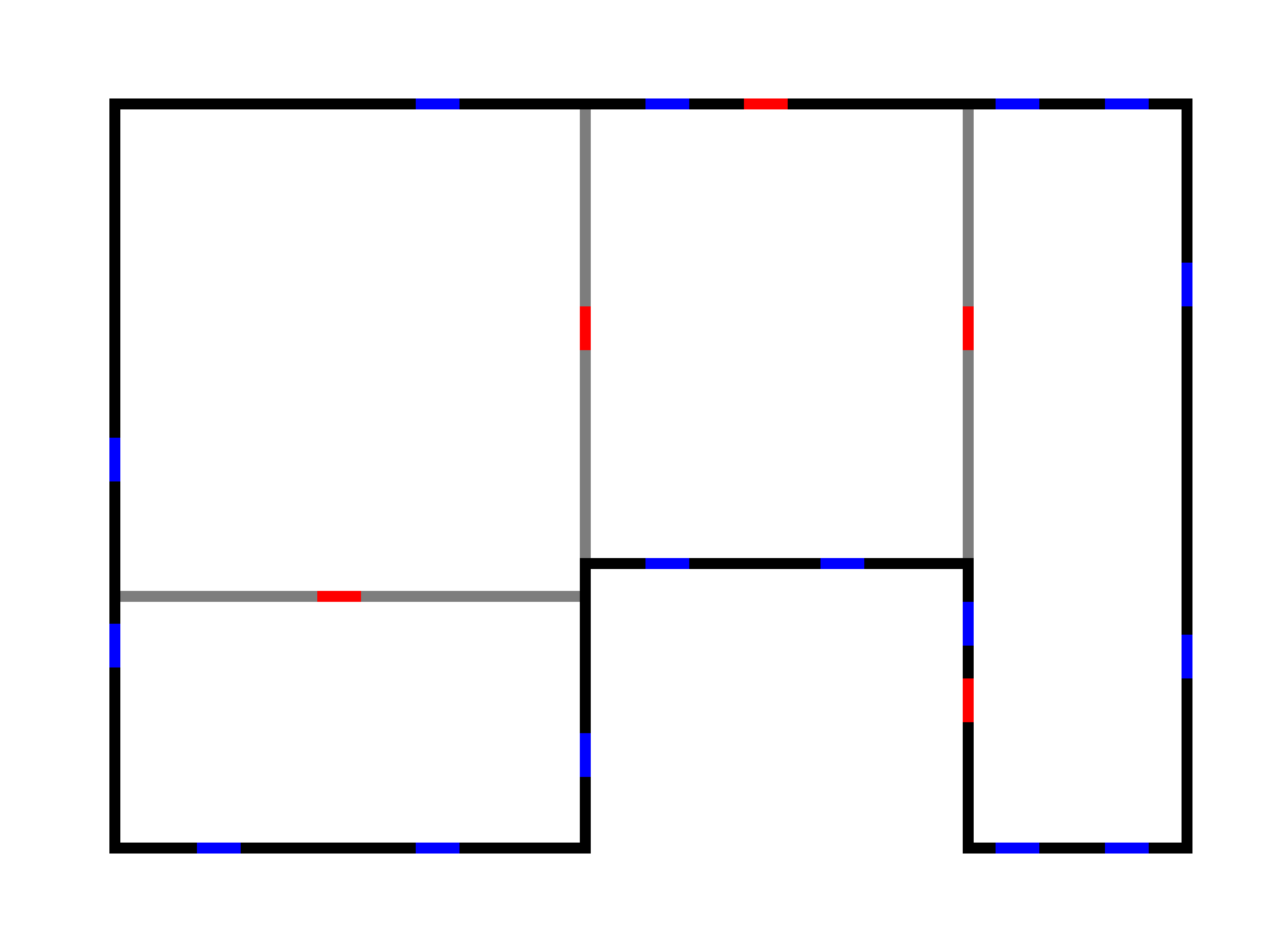} }}%
    \subfloat[$\ModelSemRoomElem$]{{\includegraphics[width=0.24\columnwidth]{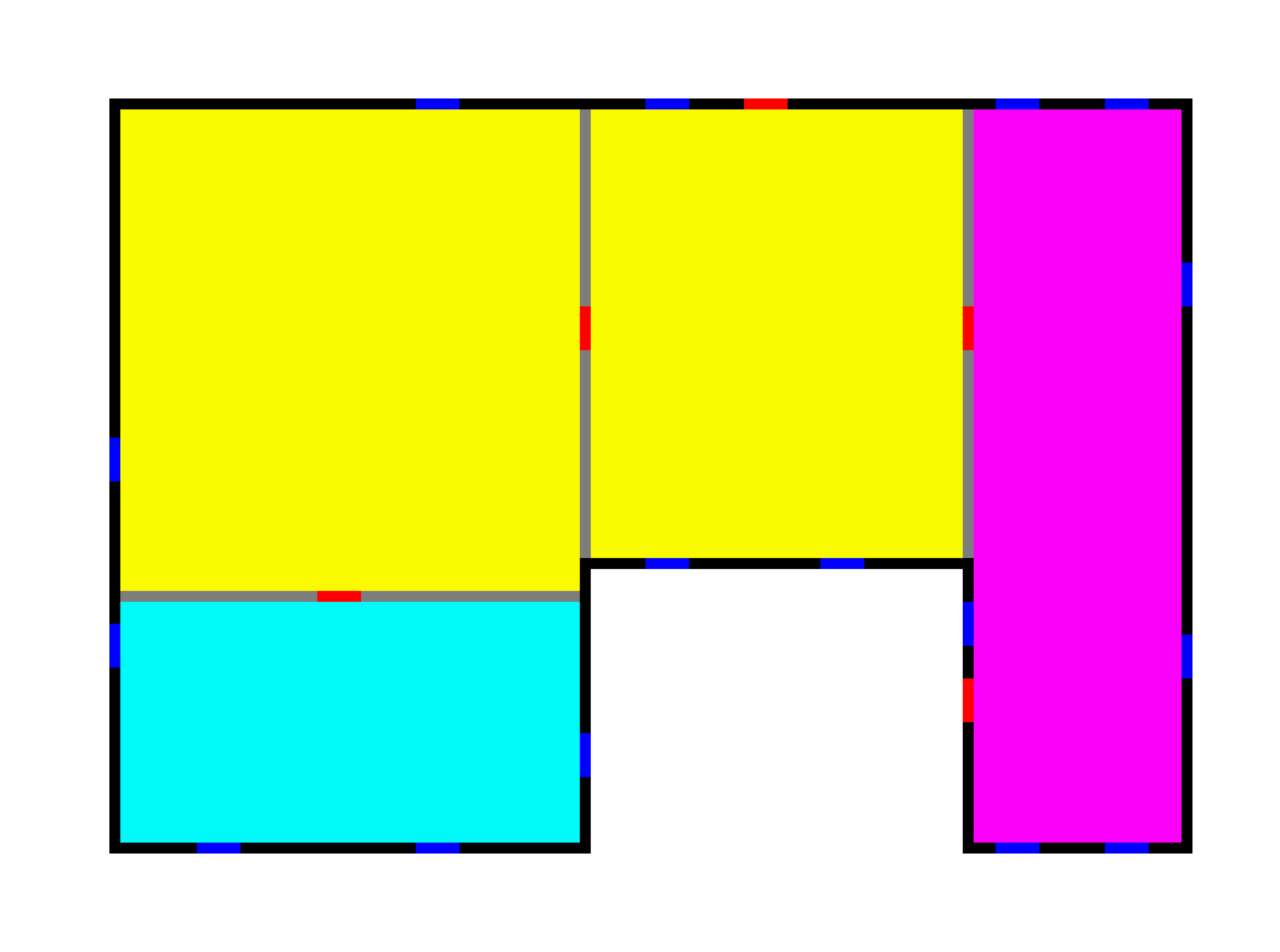} }}%
    \caption{An example floor plan created by the procedural generation algorithm, rendered in each of the four encoding styles. Images like these are combined with a description prompt to be used as the training data for the four fine-tuning approaches.}%
    \label{fig:training_data}%
\end{figure}

To further improve the fine-tuning of the diffusion models, we propose a new visual encoding approach of semantic information in floor plans. Our experiments will show  that one of the main reasons why diffusion models struggle with floor plans is the missing semantic understanding of the elements in the image. While a diffusion model may learn that the term "floor plan" is correlated to rectangular lines, it does not really understand that these lines are walls enclosing rooms. The moment we add more symbols to the image like doors, windows, furniture or assets, the model will struggle understanding the meaning of these symbols and mix them up as visible in Fig.\ \ref{fig:example_floor_plan}.
The question is: Can we make it easier for the diffusion model to learn these semantics? 

To achieve this, we propose a novel approach to encode these semantics in floor plans to improve the fine-tuning of diffusion models by adopting the following two strategy in addition to \textbf{Simplify}:
\begin{description}
\item[Encode:] When we look at a floor plan, we know that thick lines are walls and the enclosed white space is a room. We also understand its semantic function (kitchen, bedroom, living room) from the symbols of appliances that it contains (oven/fridge, bed, couch). When we remove those symbols---to simplify the model---the diffusion model (and we) cannot gain understanding about these semantics from the image. Thus, we need a way to re-encode this information, for example by filling the rooms with different colours, like cyan for bathrooms, yellow for bed/living rooms, or magenta for kitchens. We call this semantic colouring of rooms $\ModelSemRoom$.
\item[Contrast:] Floor plans are often black and white (BW) and diffusion models are designed to encode colour images. Thus, we lose lots of information capacity by just training them on black and white images. We already used colours to encode room semantics in the last step. The question then is, whether we can further simplify the floor plans by replacing the symbols for doors and windows with colours instead. We call this semantic colouring of elements $\ModelSemElem$. The combined semantic encoding of room and elements is the $\ModelSemRoomElem$ model.
\end{description}

\section{Model Training}\label{sec:finetuning}
\subsection{Generation of the Training Data }

In order to fine-tune Stable Diffusion to the different semantic encodings that we described, we implemented an algorithmic design heuristic in Python, which procedurally generates a desired number of floor plans, encodes them in each of our four different proposed styles $\ModelReduced$, $\ModelSemRoom$, $\ModelSemElem$, and $\ModelSemRoomElem$, and then generates a customized description prompt for each variant. This ensures that all four models are on even ground during the training phase. Fig.~\ref{fig:training_data} shows procedurally generated example images for each of the proposed encodings.

\begin{figure*}[t]%
    \centering
    \includegraphics[width=\textwidth]{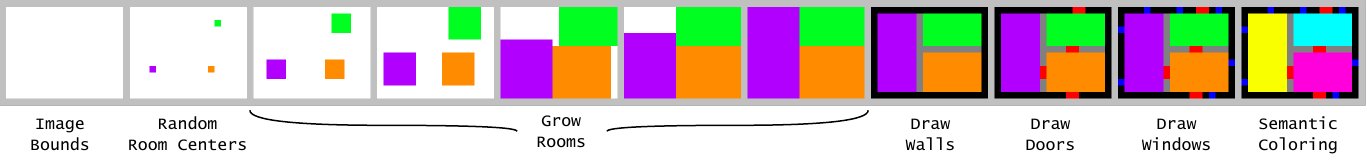}
    \caption{A low-resolution example for procedural generation in style $\ModelSemRoomElem$.}%
    \label{fig:proc_gen_example}
\end{figure*}

The floor-plan generation algorithm is based on random room locations that are grown into rectangular regions until their edges collide with the edges and corners of other regions. A visual representation of the algorithm for one of the styles is shown in Fig.~\ref{fig:proc_gen_example} and a description of the entire algorithm in simplified pseudocode is shown in Algorithm~\ref{alg:proc_gen}. Line 1--6 and images 1 and 2 show the setting of image bounds and the random selection of initial room coordinates. The process of growing the rooms is given in line 8--14 and images 3--7. The rooms grow in each direction (left $R_\I{L}$, right $R_\I{L}$ , top $R_\I{L}$ , bottom $R_\I{L}$) until they either intersect with another room or the image boundary. In lines 16--24 and images 8--11 we add walls, doors, and windows. Doors and windows are placed according to the simple rule that outside walls will often have a door flanked by two windows, while interior walls will only have one door. However, for every step and every element there is a chance for random alterations, such as the deletion of rooms or walls to create non-rectangular building arrangements. During this process we also render one variant of the plan for each style. Once the final plans are created, they are automatically checked for obvious errors and either fixed or discarded if necessary. This creates a collection of valid plans, rendered in the four different styles and labelled automatically. This final step is shown in lines 25--27. Afterwards, there is also a manual process of discarding any erroneous plans that the automated validation step did not catch, and of labelling additional aspects that would be challenging to automatically recognize, such as the general shape of the building (e.g. l-shape, o-shape, etc.). The code is publicly available on github\footnote{\url{https://github.com/AI4SC/bim-diffusion-models}}.

\begin{algorithm}
\algnewcommand\algorithmicforeach{\textbf{for each}}
\algdef{S}[FOR]{ForEach}[1]{\algorithmicforeach\ #1\ \algorithmicdo}
\begin{algorithmic}[1]
\Procedure{Procedural\_Generation}{$N_G, N_R, X_\I{max}, Y_\I{max}$}
    \For{$g = 0 \ldots N_G$}
        \State $B_\I{x} = \fn{rand}{X_\I{max}}$; $B_\I{y} = \fn{rand}{Y_\I{max}}$ \Comment{Compute random image boundary}
        \For{$i = 0 \ldots N_R$} \Comment{For all rooms $\mathfrak{R}$ with $|\mathfrak{R}|=N_R$}
            \State $R^i_\I{x} = \fn{rand}{B_\I{x}}$; $R^i_\I{y} = \fn{rand}{B_\I{y}}$ \Comment{Compute random room centers}
            \State $R^i_\I{RT} = \fn{rand}{\{\text{room},\text{kitchen},\text{bath}\}}$  \Comment{Assign random room type}
        \EndFor
        \Repeat \Comment{Repeat until no room can grow anymore}
            \State no\_update = \textit{true} \Comment{Init stop condition}
            \ForEach{$R^i \in \mathfrak{R}$}                     \Comment{For all rooms}
                \If{$R^i_\I{R}+R^j_\I{L}<R^i_\I{x}-R^j_\I{x}  \quad\forall R^j \in \mathfrak{R} | R^j_\I{x}<R^i_\I{x}$}    \Comment{No intersection with rooms right of $i$}
                    \If{$R_\I{x}+R^i_\I{R}\leq B_\I{x} $ }  \Comment{No intersection with $X$ boundary}
                        \State $R^i_\I{R} = R^i_\I{R} + 1$        \Comment{Increment room size right of center in $X$}
                        \State no\_update = \textit{false}    \Comment{Reset stop condition}
                    \EndIf
                \EndIf
                \State \vdots \Comment{Similarly increment room size left $R^i_\I{L}$, top $R^i_\I{T}$, bottom $R^i_\I{B}$ considering axis neighbours and bounds}
            \EndFor
        \Until{no\_update}
        \ForEach{$\text{S} \in \mathcal{\{\ModelReduced, \ModelSemRoom, \ModelSemElem, \ModelSemRoomElem\}}$} \Comment{Interate trough all style variants}
            \ForEach{$R \in \mathfrak{R}$}
                \State $\mathfrak{W}_R = \fn{Draw\_Walls}{R, S}$           \Comment{Draw walls in style $S$}
                \ForEach{$W \in \mathfrak{W}_R$}
                    \State $\mathfrak{D}_W = \fn{Draw\_Doors}{W, S}$       \Comment{Draw doors in style $S$}
                    \If{$\fn{Is\_Outer\_Wall}{W}$}                                \Comment{If outside facing wall}
                        \State $\mathfrak{N}_W = \fn{Draw\_Windows}{W, S}$ \Comment{Draw windows in style $S$}
                    \EndIf
                \EndFor
                \If{$\text{S} \in \mathcal{\{\ModelSemRoom, \ModelSemRoomElem\}}$}
                    \State $R_\I{c}=\fn{Color\_Room}{S, R}$               \Comment{Semantic color room in style $S$}
                \EndIf
            \EndFor
            \State $\fn{Validate\_Plan}{}$
            \State $\fn{Output\_Image}{}$
            \State $\fn{Output\_Prompt}{}$
        \EndFor
    \EndFor
\EndProcedure
\end{algorithmic}
\caption{Algorithm for the procedural generation of floor plans in four styles with $N_G$ sample number, $N_R$ number of rooms, maximum image width $X_\I{max}$ and height $Y_\I{max}$}
\label{alg:proc_gen}
\end{algorithm}

\begin{figure*}[htb]%
    \centering
    \vspace{-3pt}
    \includegraphics[width=0.8\textwidth]{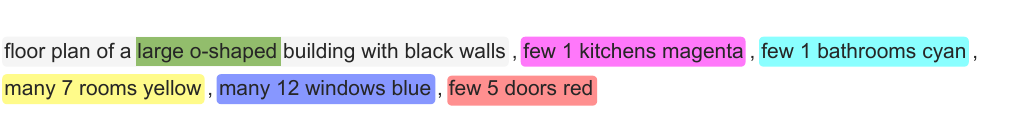}
    \caption{Example prompt used in training and generation: Green - building shape; Magenta - kitchen; Cyan - bathroom; Yellow - living rooms, Blue - windows; Red - doors}%
    \label{fig:example_prompt}
\end{figure*}

The generated prompts describe the generated floor plan and follows the common guidelines for Stable Diffusion prompts, specifically for automatic1111: Individual concepts are separated by commas and round brackets. Commas act as conceptual separators and brackets assign prompt weights to each of the concepts (always kept at 1 here). There is also a style descriptor that is exactly one token long and needs to be included at the beginning of the prompt, in order to invoke the trained style from the textual inversion. Each concept descriptor contains a number, a quantity descriptor (\textit{few} or \textit{many}), the word associated with the element itself, and a colour, as shown in Fig.~\ref{fig:example_prompt}.

\subsection{Structure of the Training Data}

The training data was created to cover various possible floor plan designs from different building shapes up to buildings that have special arrangements, like having no windows or a specific number of doors. The individual symbologies (like doors or living rooms) are also trained through manually created images that only feature these symbols in several orientations.

The following training images were generated for each style:
\begin{itemize}
    \item Two images each of individual doors, windows, three images each of living rooms, bathrooms and kitchens
    \item Ten floor plans each in which an element has an altered colour specification (green)
    \item Ten floor plans each for different building shapes: l-shaped, o-shaped, c-shaped, square, rectangle, multiple buildings
    \item Ten floor plans each for negations, i.\,e.\ elements whose count is 0 and quantity descriptors is \enquote{no}
    \item Ten floor plans each for the few and many cases of each kind of element
    \item Ten floor plans each in which the count of an element is set to 2, 4 or 6
\end{itemize}

\subsection{Fine-Tuning the Models}

We fine-tuned a Stable Diffusion model v2.1 for each encoding $\ModelReduced$, $\ModelSemRoom$, $\ModelSemElem$, and $\ModelSemRoomElem$ using Textual Inversions \cite{gal2022textual} as preliminary trials showed that it offers fastest training time at equal result performance for our use case in comparison to \cite{hu2021lora} and \cite{Ruiz_2023_CVPR}.

With each batch of generated training data, we fine-tuned a Stable Diffusion model v2.1 for each encoding $\ModelReduced$, $\ModelSemRoom$, $\ModelSemElem$, and $\ModelSemRoomElem$ using Textual Inversions \cite{gal2022textual}, as preliminary explained in Section \ref{sec:tuning}. Preliminary trials showed that it offers fastest training time at equal result performance for our use case in comparison to \cite{hu2021lora} and \cite{Ruiz_2023_CVPR}. We utilized the \textit{automatic1111} platform with the DreamBooth extension to run the fine-tuning. The source model was the v2-1\_768-ema-pruned checkpoint by StabilityAI. We used the recommended training parameters for tuning a new style for this model, with 100 training epochs, a 768x768 maximum resolution and a one-token instance prompt. The latter creates a style descriptor that we then use to specifically invoke our fine-tuning during image generation.

\begin{figure}[t]%
    \centering
    \subfloat[$\ModelReduced$]{{\includegraphics[width=0.24\columnwidth]{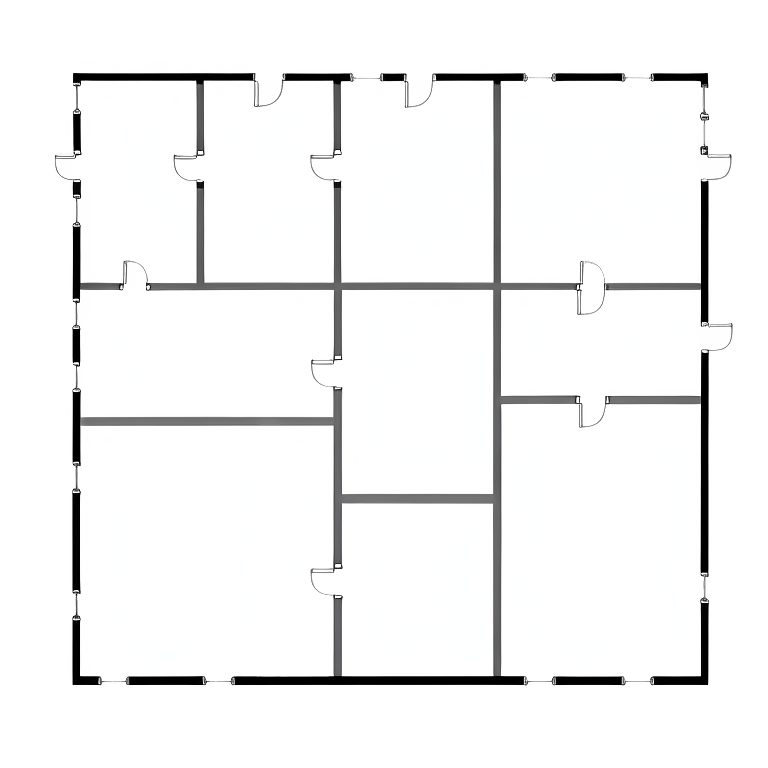} }}%
    \subfloat[$\ModelSemRoom$]{{\includegraphics[width=0.24\columnwidth]{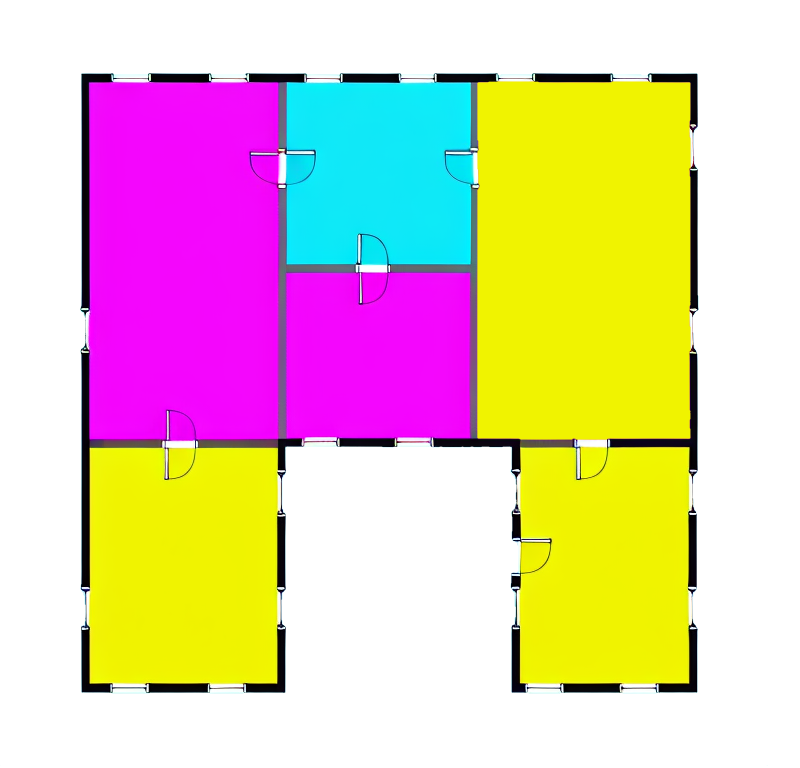} }}
    \subfloat[$\ModelSemElem$]{{\includegraphics[width=0.24\columnwidth]{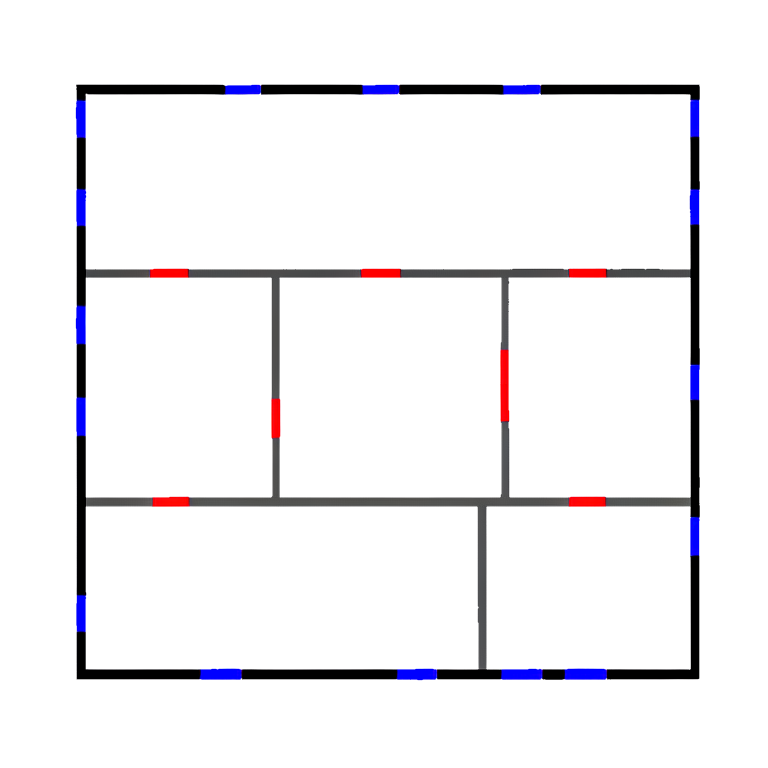} }}%
    \subfloat[$\ModelSemRoomElem$]{{\includegraphics[width=0.24\columnwidth]{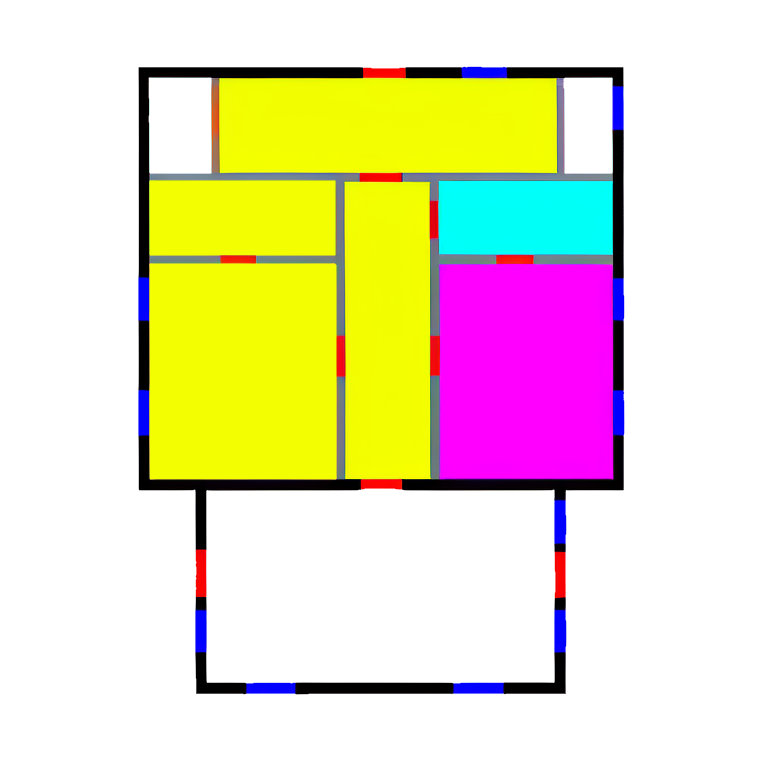} }}%
    \caption{Successful example generations for each of the four retrained models.}%
    \label{fig:generations}%
\end{figure}

We created a model variant for each style. Every model was able to attain reasonable similarity to the trained style for simple prompts, as shown in Fig.~\ref{fig:generations}. We will evaluate the precise performance of each model for our use case through quantitative quality metrics in the next section.

\section{Evaluation}\label{sec:experiments}

\subsection{Methodology of Validation}

To compare the performance of the different un-tuned and fined-tuned diffusion models we defined a large set of 43 test prompts that represent various practical request as well as specific challenging situations for the diffusion models. These prompts are based on the structure of the training prompts, and were created such that they each invoke different aspects of that training. The prompts are classified into the following categories, according to which challenges they pose to the model:
\begin{description}
\item[Valid Plan:] First, we evaluate how many of the generated images show a valid floor plan that is not cut off and contains at least some recognizable symbology. 
\item[Overfit Check:] Second, we check if the fine-tuned models do not overfit on the training samples. We consider a model overfitted if it responds to a prompt asking for a face with a floor plan, as “face” is semantically far from floor plan, we assume that the model always responds with floor plans even if not asked for and thus does not differentiate the tokens well enough. 
\item[Quantify Objects/Rooms:] To see if the model can understand quantities we query for \enquote{few} (less than 7) and \enquote{many} (more than 6) windows/doors and rooms/kitchens/bathrooms.
\item[Count of Objects/Rooms:] Instead of asking for fuzzy quantities, we ask for specific counts (2, 4, 6) of windows/doors and rooms/kittchen/bathrooms.
\item[Remove Objects/Rooms:] To check if the model has semantic understanding of core terms, we ask it to remove them and e.\,g.\ create a building with no windows/doors or no kitchens, etc. (Note: this is not done through negative prompts, as we want to test the ability of the models to turn structured queries into semantically sound plans.)
\item[Recolour Objects/Rooms:] To see if the model has semantic understanding of the colour and term encoding that we use, we ask special queries to change the colour of doors/windows and rooms like kitchen, etc.
\item[Arrange Rooms:] To see if the model has spatial understanding, we request specific shapes of the room arrangement like L-, C-, or O-shapes, rectangular shapes, and multiple separated buildings.
\end{description}

For each test prompt and diffusion model in $\ModelBaseline$, $\ModelReduced$, $\ModelSemRoom$, $\ModelSemElem$, $\ModelSemRoomElem$ we generate 10 sample images. To further improve results of all generations, we also added the following negative prompt to each iteration: "gradients, blurry, fuzzy borders, reflections, lighting". For each sample image we evaluated manually whether or not (1 or 0) the feature requested in the prompt was addressed in the generated image. This process resulted in 2.150 sample images that we manually evaluated.


\subsection{Results}

Table~\ref{tab:results} shows the results of our evaluation. The baseline out-of-the-box diffusion model $\ModelBaseline$ without fine tuning performs the worst. Only 6\,\% of all generated floor plans were fully valid, while most floor plans were faulty in some aspect. Examples for faults are rooms that are missing walls, symbols that are partially unrecognizable or room types that are not clear (see Fig.~\ref{fig:bad}). 

\begin{table}[tbh]
\centering
\caption{Quantitative results of the experimental validation of the four trained models with $\#$ Number of Experiments} 
\label{tab:results}
\begin{tabular}{l|r|rrrrr}
\toprule
                   & $\#$ & $\ModelBaseline$ & $\ModelReduced$   & $\ModelSemRoom$ & $\ModelSemElem$ & $\ModelSemRoomElem$ \\
\midrule 
Valid Plan         & 410 & 6\,\%  & 63\,\%  & 57\,\%    & \textbf{90\,\%}    & 66\,\%      \\
Not Overfitted     &  20 & 80\,\%  & 95\,\% & 80\,\%   & 95\,\%   & \textbf{100\,\%}     \\
\hline
Quantify Obj.  &  40 & 65\,\%  & \textbf{68\,\%} & 50\,\%   & 55\,\%   & 55\,\%     \\
Quantify Rooms      &  80 & 19\,\%  & 15\,\%  & 58\,\%  & \textbf{65\,\%}    & 45\,\%     \\
\hline
Count of Obj.      &  60 & 3\,\%  & \textbf{5\,\%} & 2\,\%    & 2\,\%    & 0\,\%      \\
Count of Rooms     & 120 & 10\,\%  & 10\,\%  & \textbf{12\,\%}  & 7\,\%    & 10\,\%      \\
\hline
Remove Obj.        &  20 & 15\,\%  & 0\,\%  & 20\,\%   & \textbf{40\,\%}   & 30\,\%     \\
Remove Room        &  30 & 0\,\%  & 0\,\% & 10\,\%   & 0\,\%    & \textbf{13\,\%}    \\
\hline
Recoloured Obj.     &  30 & \textbf{3\,\%}  & 0\,\%  & 0\,\%    & 0\,\%    & 0\,\%      \\
Recolour Room       &  30 & 0\,\%  & 0\,\%  & \textbf{3\,\%}    & 0\,\%    & 0\,\%      \\
\hline
Arrange Rooms      &  50 & 24\,\%  & \textbf{62\,\%} & 34\,\%   & 56\,\%   & 30\,\%     \\
\bottomrule
\end{tabular}
\end{table}

Based on these results, Stable Diffusion is unable to generate useful floor plans before additional fine-tuning. Once finetuned these capabilities drastically change, as visible in the \textit{valid plan} category. The tuned models mostly manage to display a full building instead of partial cut-outs. There is also recurring symbology that avoids visual artifacts and errors that render a plan difficult to interpret or even entirely unreadable. The model with the best performance here is $\ModelSemElem$. It is the most visually simple model, without room colouring or complex door and window symbols. Demonstrating that the \textbf{Simplify} strategy helps generating valid models.

The \textit{overfitting} test results show that the fine-tuned models are not overfitted. Even the baseline models do not always process the overfitting query correctly and all tuned models are in the same margin of error.

Model $\ModelSemElem$ was also the model that responds best to requests to \textit{removal of} specific elements. Most other models perform badly in this category. This shows that our semantic \textbf{Encoding} strategy helps the diffusion model learn what windows and doors are. That learning this encoding is not strictly associated with just colour is evidenced by the results of the \textit{recolouring} prompts, which usually failed.

Similarly, the results of the \textit{count} prompts show that the models simply are not yet capable of counting, even after fine-tuning models with many examples with correct counts. The results for \textit{quantify} prompts are significantly better. This may partially be because we only have two categories (few/many), resulting in a high likelihood for false positives. The baseline model $\ModelBaseline$ for example usually creates incomplete cut-off plans with few elements, and thus gets all few windows/doors/rooms queries correct. Yet, the semantic encoding model $\ModelSemElem$ seems to again perform above the random draw chance.

The prompts testing room \textit{arrangements} like L-, C- or O-shapes work well enough across all models, at least in comparison to many other queries. This shows that even the baseline model has some basic understanding of arrangements in the higher diffusion model layers. This still is improved significantly with fine-tuning. Here the models $\ModelReduced$ and $\ModelSemElem$ show the highest performance increase. This is likely because the walls and rooms are very uniform in colour, increasing the contrast between inside and outside, which is harder to differentiate for the multicoloured rooms. This result partially supports our \textbf{Contrasting} strategy, although it shows that the three strategies can sometimes run counter to each other and thus require careful balancing.

\subsection{Discussion}

The results from the experiments show that fine-tuning diffusion models significantly improves validity of the generated floor plans from 6\,\% of the untuned model to 57\,\%--90\,\% of the fine-tuned models. Tuning the models directly with semantic element encoding $\ModelSemElem$ is resulting in the best performance across all test prompts. This is proving the correctness of our three strategies \textbf{Simplify}, \textbf{Encode} and \textbf{Contrast}, as this model has the best trade-off between them. It is \textit{simple} in that it contains only lines. By semantically \textit{encoding} doors and windows with colours we keep them in the model without reducing simplicity and creating images with high \textit{contrast} that feature only black, white, grey and two colours. Adding room colours reduces this contrast, which is why the $\ModelSemRoom$ and $\ModelSemRoomElem$ models are performing worse.

\begin{figure}[t]%
    \centering
    \subfloat[]{{\includegraphics[width=0.30\columnwidth]{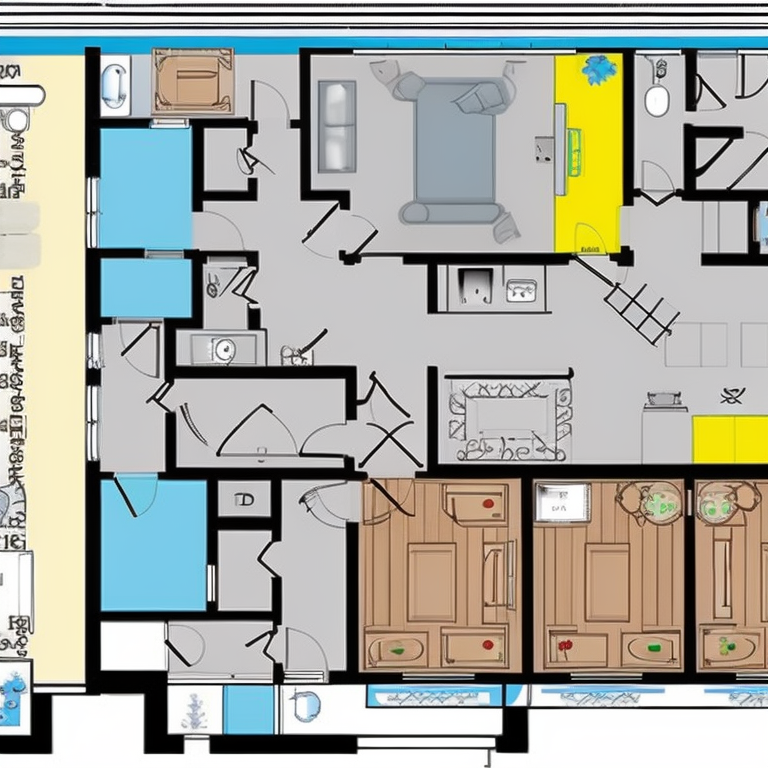} }\quad}%
    \subfloat[]{{\includegraphics[width=0.30\columnwidth]{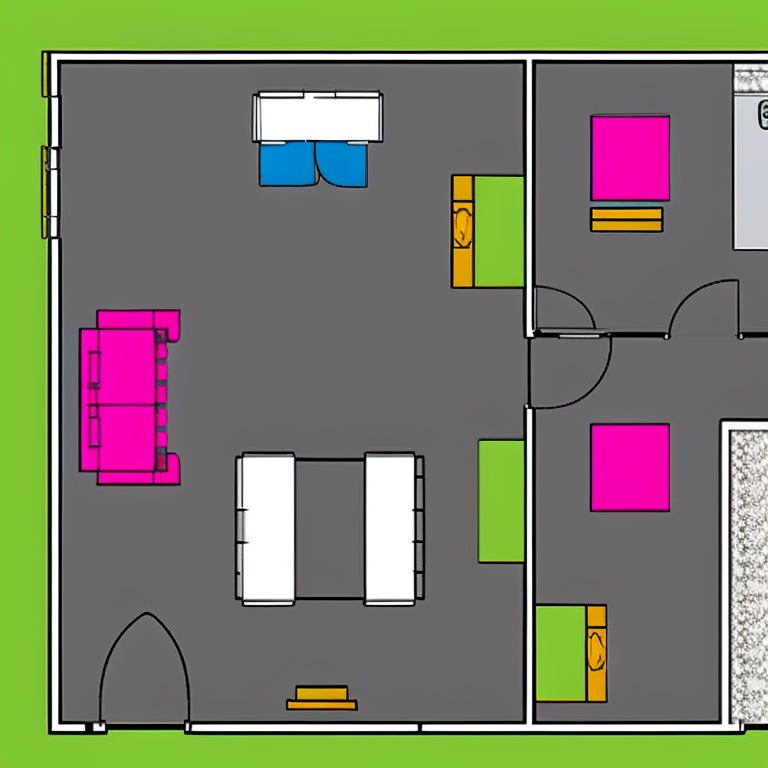} }\quad}%
    \subfloat[]{{\includegraphics[width=0.30\columnwidth]{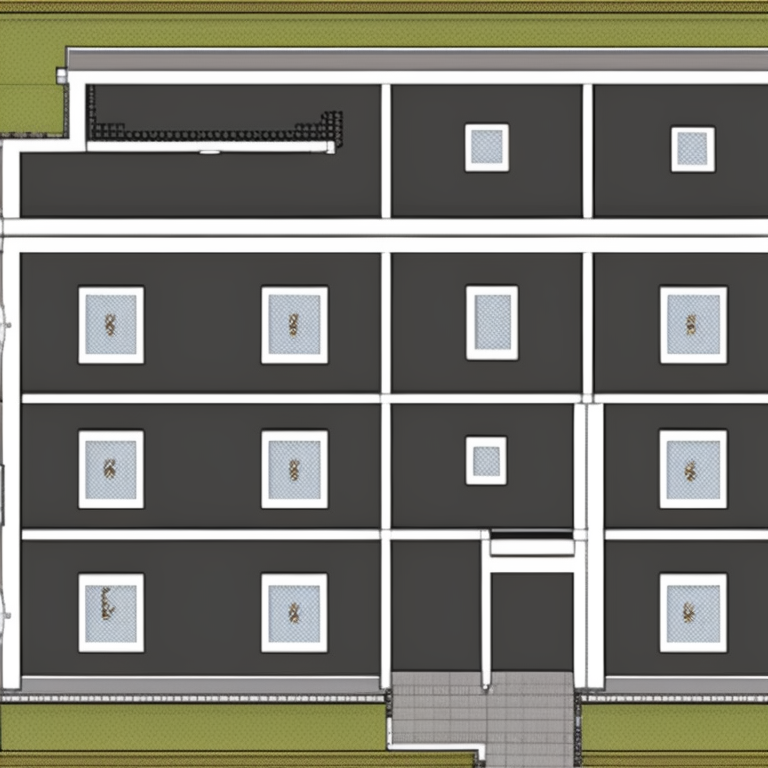} }}%
    \caption{Examples of generation errors: a) overlapping door-like symbols; b) incorrect symbols and colours; c) building front-view instead of floor-plan encodings of the generated images. Approaches within the area of diffusion models are restoration models, that for example can repair faces in generated images \cite{wang2021gfpgan}}%
    \label{fig:bad}%
\end{figure}

Our results also show that diffusion models still have major limitations in generating floor plans to satisfy queried requirements. The best performing queries for quantities do not exceed a 68\,\% success rate. Queries around counts and recolouring usually fail completely. Examples of common errors in the context of our evaluation are shown in Fig.~\ref{fig:bad}. They can be summarized as:
\begin{itemize}
    \item Results show only partial floor plans.
    \item Requested elements are not contained in the generated result.
    \item Symbols and colour palettes are generated in incorrect context.
    \item Contextual layout instructions are ignored (e.\,g.\ kitchen left of living room).
    \item Requested counts of elements and rooms are only met by chance.
\end{itemize}

There are of course other more general issues in the current generation of diffusion models. The more semantically detailed we want to get, the more difficult it would be to train models, like for example for the handling of different door and window sub-types and their correct symbols. Many models also have inherent quality limitations, for example regarding image size. Models can also reinforce bad design principles, as the aesthetics of image compositions that are encoded within them do not necessarily correspond to good architecture. These issues point to many important future research directions that we will identify in the following section.

\section{Future Directions}
\label{sec:results}

\subsection{Image-based Diffusion Models}
Our experiments revealed several limitations of diffusion models in generating floor plans. The success rates of prompts in our experiments are far from optimal. This is one of the reasons why diffusion models usually generate multiple samples at once and let the user choose. The generation itself happens within seconds, significantly outperforming evolutionary generative design approaches. The analysis of Midjourney queries in \cite{ploennigs2023ai} showed that only 13.1\,\% of all prompts are upscaled and then refined over 6 iterations---which does not seem to be an issue for the popularity of the platform. Good \textit{workflows that allow users to quickly select and refine queries} for computational design are important, to enable quick filtering of bad samples as shown in our idealized workflow in Fig.\ \ref{fig:ideal_workflow}.

Furthermore, the huge body of ongoing research in diffusion models will lead to improved models soon. An important topic in this fundamental area is their \textit{capability to correctly count elements} in the generated images \cite{Yang_2023_CVPR}. But, this still requires that the diffusion model has a good understanding of the semantics of the elements in a floor plan. If it does not know what symbol represents a door it will not be able to count it correctly.


This lack of explicit semantic understanding is a main limitation of diffusion models. Our experiments showed that it is possible to tune diffusion models to improve the results, but it is not currently feasible to remove all limitations and requires further \textit{research in semantic diffusion model encodings}. Related to this is the question of how we can turn visual semantic encodings like we used in this paper back into valid plan symbols. Possible approaches within the area of diffusion models are \textit{restoration models}, that for example can repair faces in generated images \cite{wang2021gfpgan}. The models have a similar lack of semantic understanding about additional context, like the surrounding buildings or site planning. While our results in specifying specific shapes were promising, we did not evaluate the external environment.

There is also the open research question of \textit{transforming generated binary images to Building Information Models}, which we identified as the ideal result of generative workflows in Fig.\ \ref{fig:ideal_workflow}. The benefit of our semantic encoding is that our strategies of Simplify, Encode, and Contrast also create cleaner images for BIM restoration approaches like \cite{pizarro2022automatic}.

\subsection{BIM-based Diffusion Models }
All the future research topics mentioned so far address the limitations of image-based diffusion models in computational design. However, the ideal domain-specific solution would be models that directly operate on BIM data. This leaves us with the final essential question: Why not develop a diffusion model for BIM models and thus strip out any information loss incurred by data conversions in and out of the process?

From our introduction we can derive some requirements for such a BIM-based diffusion model:
\begin{itemize}
    \item The model needs to be \textbf{multi-scale}. Traditionally diffusion models work on different resolutions to repair an image successively. These layers then store different information of the original image.
    \item The model needs to be \textbf{diffusable}. We need to be able to ingest noise into the model on each level to be able to train a repair network.
    \item The model needs to be \textbf{semantically labelled}. We need to be able to label objects and properties as additional input to the diffusion model.
\end{itemize}

BIM models easily fulfil the last requirement. Given their nature, they are well structured and semantically rich models. Unfortunately, the first two requirements are not easily applicable to BIM models. We can of course raster a BIM model in different resolutions to generate binary images. But our results show that we lose the semantic information and the well-structured nature in the process, and that these aspects are difficult to recreate. 

Our proposal is to research \textit{BIM-based Diffusion Models} that instead keep semantic information and utilize the hierarchical nature of BIM models. In a BIM model we have floors, zones, and objects that form a hierarchy. Now the only requirement we must solve is the diffusability. For that, we need a Neural Network representation of the BIM model that we can easily alter. Here we can utilize the fact that at the core of BIM models like IFC lies a graph representation like STEP. Representing BIM models in Graphs was proposed before for generative approaches like \cite{gan2022bim,li2023automated}. 
The closest current neural network representations for such structures are Graph Neural Networks (GNN) recently developed for e.\,g.\ time series forecasting \cite{ba2022automated}. We propose research into multi-layer GNN models that represent on each layer the respective information layer of a BIM model, like which floors exists and which are close to each other, which zones exists within these floors and how they are connected via walls. Diffusion approaches for graphs capable of weakening or removing those connections in the graph were recently developed \cite{chamberlain2021grand}. Also, early approaches for using diffusion models for polygonal floor plans exist \cite{shabani2023housediffusion,gueze2023floor}, but they lack the semantic information of BIM.

\begin{figure}[t]%
    \centering
    \subfloat[\centering Training]{{\includegraphics[width=0.75\columnwidth]{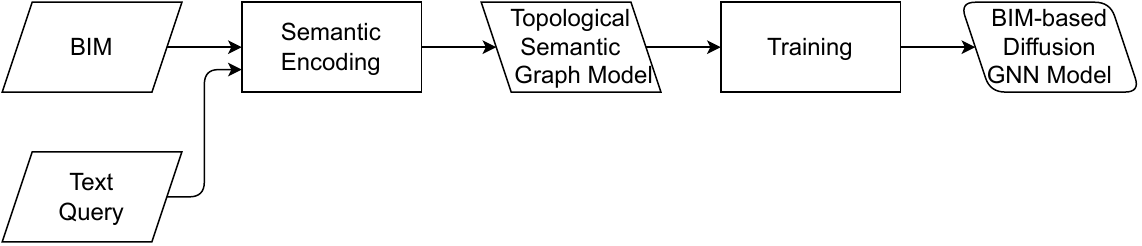} }}\\%
    \subfloat[\centering Generation]{{\includegraphics[width=0.75\columnwidth]{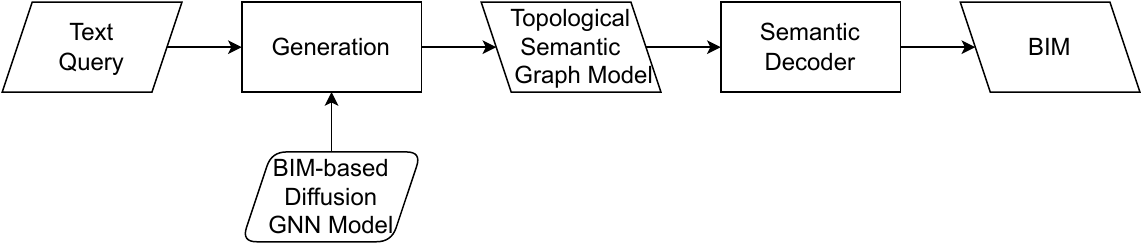} }}%
    \caption{Workflow for a Semantic BIM-based Diffusion Model}%
    \label{fig:semantic_gnn_model}%
\end{figure}

The benefit of this approach is that, utilizing the BIM structure, we could directly generate 3D models instead of just 2D plans. A challenge here are the data requirements for training. As discussed in the beginning, training a diffusion model usually requires huge amounts of data. Here, it may be one option to use \textit{algorithmic design approaches to generate training data} like we did for the fine-tuning in this paper. Alternatively, we as a research community may need to start an \textit{open-source initiative to collect open BIM datasets} at a large scale to enable this next level of \textit{BIM-based Diffusion Models}. How our workflow would look with this kind of model, is shown in Fig.~\ref{fig:semantic_gnn_model}.



\section{Conclusion}
In this paper we evaluated the use of diffusion models for computational design for the example of floor plans. We analysed the current state of the art in the field and explained diffusion model approaches. We identified the main limitations and developed a novel approach to improve fine-tuning of diffusion models through semantic encoding. We analysed the workflows for using the diffusion models qualitatively and evaluate the performance of different diffusion model encodings quantitatively in a large set of experiments. Architectural quality of the floor plans was not evaluated for now, and would need a separate study, for example involving expert reviews.

Our experiments show that current diffusion models can already create floor plans and that results can be improved significantly with fine-tuning. Particularly our proposed semantic encoding improved validity of the generated floor plans to 90\,\%. However, our experiments also unveiled several shortcomings in the current diffusion models, of which many track back to the lack of semantic understanding. Therefore, we identified several future directions for research in this new field of AI-driven computational design, which may in future ultimately be able to automate many detailed planning steps. It is as of yet unclear whether we can get there with just this is possible to achieve through appropriately powerful image-based diffusion models, or whether we need to start working on BIM-based Diffusion Models to unlock their semantic potential.


\section{Declarations}
\paragraph{Data Availability Statement} The code for generating the training samples as as well as all validation queries are available on \url{https://github.com/AI4SC/bim-diffusion-models}.
\paragraph{Competing interests} The authors have no financial or proprietary interests in any material discussed in this article.
\paragraph{Author contributions} JP: Ideation and manuscript, Writing Sec. 1, 2.1--2.4, 4, 5. MB: Use Case Experiments, Writing Sec. 2.5, 3. Both authors reviewed and edited all sections as well as read and approved the final manuscript.

\bibliographystyle{unsrtnat}
\bibliography{main}

\begin{thebibliography}{45}
\providecommand{\natexlab}[1]{#1}
\providecommand{\url}[1]{\texttt{#1}}
\expandafter\ifx\csname urlstyle\endcsname\relax
  \providecommand{\doi}[1]{doi: #1}\else
  \providecommand{\doi}{doi: \begingroup \urlstyle{rm}\Url}\fi

\bibitem[Ploennigs and Berger(2023)]{ploennigs2023ai}
Joern Ploennigs and Markus Berger.
\newblock {AI} art in architecture.
\newblock \emph{{AI} in Civil Engineering}, 2\penalty0 (1):\penalty0 8, 2023.

\bibitem[Ploennigs and Berger(2024)]{ploennigs2023generative}
Joern Ploennigs and Markus Berger.
\newblock \emph{Decoding Cultural Heritage with {AI}}, chapter Generative {AI}
  and the History of Architecture.
\newblock Springer, 2024.

\bibitem[Van~Tam et~al.(2023)Van~Tam, Quoc~Toan, Phong, and
  Durdyev]{van2023impact}
Nguyen Van~Tam, Nguyen Quoc~Toan, Vu~Van Phong, and Serdar Durdyev.
\newblock Impact of {BIM}-related factors affecting construction project
  performance.
\newblock \emph{Int. Journal of Building Pathology and Adaptation}, 41\penalty0
  (2):\penalty0 454--475, 2023.

\bibitem[Barbosa et~al.(2017)Barbosa, Woetzel, and
  Mischke]{barbosa2017reinventing}
Filipe Barbosa, Jonathan Woetzel, and Jan Mischke.
\newblock Reinventing construction: A route of higher productivity.
\newblock Technical report, McKinsey Global Institute, 2017.

\bibitem[Pan and Zhang(2021)]{PAN2021103517}
Yue Pan and Limao Zhang.
\newblock Roles of artificial intelligence in construction engineering and
  management: A critical review and future trends.
\newblock \emph{Automation in Construction}, 122:\penalty0 103517, 2021.
\newblock ISSN 0926-5805.

\bibitem[Pan and Zhang(2023)]{pan2023integrating}
Yue Pan and Limao Zhang.
\newblock Integrating {BIM} and {AI} for smart construction management: Current
  status and future directions.
\newblock \emph{Archives of Computational Methods in Engineering}, 30\penalty0
  (2):\penalty0 1081--1110, 2023.

\bibitem[Debrah et~al.(2022)Debrah, Chan, and Darko]{debrah2022artificial}
Caleb Debrah, Albert~PC Chan, and Amos Darko.
\newblock Artificial intelligence in green building.
\newblock \emph{Automation in Construction}, 137:\penalty0 104192, 2022.

\bibitem[Sohl-Dickstein et~al.(2015)Sohl-Dickstein, Weiss, Maheswaranathan, and
  Ganguli]{sohl2015deep}
Jascha Sohl-Dickstein, Eric Weiss, Niru Maheswaranathan, and Surya Ganguli.
\newblock Deep unsupervised learning using nonequilibrium thermodynamics.
\newblock In \emph{{ICML}}, pages 2256--2265, 2015.

\bibitem[Caetano et~al.(2020)Caetano, Santos, and
  Leit{\~a}o]{caetano2020computational}
In{\^e}s Caetano, Lu{\'\i}s Santos, and Ant{\'o}nio Leit{\~a}o.
\newblock Computational design in architecture: Defining parametric,
  generative, and algorithmic design.
\newblock \emph{Frontiers of Architectural Research}, 9\penalty0 (2):\penalty0
  287--300, 2020.

\bibitem[G{\"u}rsel~Dino(2012)]{gursel2012creative}
{\.I}pek G{\"u}rsel~Dino.
\newblock Creative design exploration by parametric generative systems in
  architecture.
\newblock \emph{METU J. Faculty of Architecture}, 2012.

\bibitem[Jabi(2013)]{jabi2013parametric}
Wassim Jabi.
\newblock \emph{Parametric design for architecture}.
\newblock Hachette UK, 2013.

\bibitem[Schnabel(2007)]{schnabel2007parametric}
Marc~Aurel Schnabel.
\newblock Parametric designing in architecture.
\newblock In \emph{Computer-aided architectural design futures (CAADFutures)},
  pages 237--250. Springer, 2007.

\bibitem[Burry(1993)]{burry1993expiatory}
Mark Burry.
\newblock \emph{Expiatory church of the sagrada familia: Antoni Gaudi}.
\newblock Phaidon Press, 1993.
\newblock ISBN 0714828491.

\bibitem[Monedero(2000)]{monedero2000parametric}
Javier Monedero.
\newblock Parametric design: a review and some experiences.
\newblock \emph{Automation in construction}, 9\penalty0 (4):\penalty0 369--377,
  2000.

\bibitem[Rempling et~al.(2019)Rempling, Mathern, Ramos, and
  Fern{\'a}ndez]{rempling2019automatic}
Rasmus Rempling, Alexandre Mathern, David~Tarazona Ramos, and Santiago~Luis
  Fern{\'a}ndez.
\newblock Automatic structural design by a set-based parametric design method.
\newblock \emph{Automation in Construction}, 108:\penalty0 102936, 2019.

\bibitem[Granadeiro et~al.(2013)Granadeiro, Duarte, Correia, and
  Leal]{granadeiro2013building}
Vasco Granadeiro, Jos{\'e}~P Duarte, Jo{\~a}o~R Correia, and V{\'\i}tor~MS
  Leal.
\newblock Building envelope shape design in early stages of the design process:
  Integrating architectural design systems and energy simulation.
\newblock \emph{Automation in construction}, 32:\penalty0 196--209, 2013.

\bibitem[Caldas(2008)]{caldas2008generation}
Luisa Caldas.
\newblock Generation of energy-efficient architecture solutions applying
  {GENE\_ARCH}: An evolution-based generative design system.
\newblock \emph{Advanced Engineering Informatics}, 22\penalty0 (1):\penalty0
  59--70, 2008.

\bibitem[Krish(2011)]{krish2011practical}
Sivam Krish.
\newblock A practical generative design method.
\newblock \emph{Computer-Aided Design}, 43\penalty0 (1):\penalty0 88--100,
  2011.

\bibitem[Chaszar and Joyce(2016)]{chaszar2016generating}
Andre Chaszar and Sam~Conrad Joyce.
\newblock Generating freedom: Questions of flexibility in digital design and
  architectural computation.
\newblock \emph{Int. J.of Architectural Computing}, 14\penalty0 (2):\penalty0
  167--181, 2016.

\bibitem[Oxman(2017)]{oxman2017thinking}
Rivka Oxman.
\newblock Thinking difference: Theories and models of parametric design
  thinking.
\newblock \emph{Design studies}, 52:\penalty0 4--39, 2017.

\bibitem[Caetano and Leit{\~a}o(2019)]{caetano2019integration}
In{\^e}s Caetano and Ant{\'o}nio Leit{\~a}o.
\newblock Integration of an algorithmic {BIM} approach in a traditional
  architecture studio.
\newblock \emph{J.of Computational Design and Engineering}, 6\penalty0
  (3):\penalty0 327--336, 2019.

\bibitem[Sun et~al.(2022)Sun, Wu, Liu, Min, Zhang, and Zheng]{sun2022wallplan}
Jiahui Sun, Wenming Wu, Ligang Liu, Wenjie Min, Gaofeng Zhang, and Liping
  Zheng.
\newblock Wallplan: synthesizing floorplans by learning to generate wall
  graphs.
\newblock \emph{ACM Transactions on Graphics (TOG)}, 41\penalty0 (4):\penalty0
  1--14, 2022.

\bibitem[Mandow et~al.(2020)Mandow, P{\'e}rez-de-la Cruz,
  Rodr{\'\i}guez-Gavil{\'a}n, and Ruiz-Montiel]{mandow2020architectural}
Lawrence Mandow, Jos{\'e}-Luis P{\'e}rez-de-la Cruz, Ana~Bel{\'e}n
  Rodr{\'\i}guez-Gavil{\'a}n, and Manuela Ruiz-Montiel.
\newblock Architectural planning with shape grammars and reinforcement
  learning: Habitability and energy efficiency.
\newblock \emph{Engineering Applications of Artificial Intelligence},
  96:\penalty0 103909, 2020.

\bibitem[Weber et~al.(2022)Weber, Mueller, and Reinhart]{weber2022automated}
Ramon~Elias Weber, Caitlin Mueller, and Christoph Reinhart.
\newblock Automated floorplan generation in architectural design: A review of
  methods and applications.
\newblock \emph{Automation in Construction}, 140:\penalty0 104385, 2022.

\bibitem[Chang et~al.(2021)Chang, Cheng, Luo, Murata, Nourbakhsh, and
  Tsuji]{chang2021building}
Kai-Hung Chang, Chin-Yi Cheng, Jieliang Luo, Shingo Murata, Mehdi Nourbakhsh,
  and Yoshito Tsuji.
\newblock {Building-GAN}: Graph-conditioned architectural volumetric design
  generation.
\newblock In \emph{IEEE/CVF Int. Conf. on Computer Vision}, pages 11956--11965,
  2021.

\bibitem[Wu et~al.(2022)Wu, Stouffs, and Biljecki]{wu2022generative}
Abraham~Noah Wu, Rudi Stouffs, and Filip Biljecki.
\newblock Generative adversarial networks in the built environment: A
  comprehensive review of the application of gans across data types and scales.
\newblock \emph{Building and Environment}, page 109477, 2022.

\bibitem[Nauata et~al.(2021)Nauata, Hosseini, Chang, Chu, Cheng, and
  Furukawa]{nauata2021house}
Nelson Nauata, Sepidehsadat Hosseini, Kai-Hung Chang, Hang Chu, Chin-Yi Cheng,
  and Yasutaka Furukawa.
\newblock {House-gan++}: Generative adversarial layout refinement network
  towards intelligent computational agent for professional architects.
\newblock In \emph{IEEE/CVF Conf. on Computer Vision and Pattern Recognition},
  pages 13632--13641, 2021.

\bibitem[Ho et~al.(2020)Ho, Jain, and Abbeel]{ho2020denoising}
Jonathan Ho, Ajay Jain, and Pieter Abbeel.
\newblock Denoising diffusion probabilistic models.
\newblock In \emph{NeurIPS}, volume~33, pages 6840--6851, 2020.

\bibitem[Radford et~al.(2021)Radford, Kim, Hallacy, Ramesh, Goh, Agarwal,
  Sastry, Askell, Mishkin, Clark, et~al.]{radford2021learning}
Alec Radford, Jong~Wook Kim, Chris Hallacy, Aditya Ramesh, Gabriel Goh,
  Sandhini Agarwal, Girish Sastry, Amanda Askell, Pamela Mishkin, Jack Clark,
  et~al.
\newblock Learning transferable visual models from natural language
  supervision.
\newblock In \emph{{ICML}}, pages 8748--8763, 2021.

\bibitem[Ramesh et~al.(2021)Ramesh, Pavlov, Goh, Gray, Voss, Radford, Chen, and
  Sutskever]{ramesh2021zero}
Aditya Ramesh, Mikhail Pavlov, Gabriel Goh, Scott Gray, Chelsea Voss, Alec
  Radford, Mark Chen, and Ilya Sutskever.
\newblock Zero-shot text-to-image generation.
\newblock In \emph{{ICML}}, pages 8821--8831, 2021.

\bibitem[Ruiz et~al.(2023)Ruiz, Li, Jampani, Pritch, Rubinstein, and
  Aberman]{Ruiz_2023_CVPR}
Nataniel Ruiz, Yuanzhen Li, Varun Jampani, Yael Pritch, Michael Rubinstein, and
  Kfir Aberman.
\newblock Dreambooth: Fine tuning text-to-image diffusion models for
  subject-driven generation.
\newblock In \emph{{CVPR}}, pages 22500--22510, June 2023.

\bibitem[Hu et~al.(2021)Hu, Shen, Wallis, Allen-Zhu, Li, Wang, Wang, and
  Chen]{hu2021lora}
Edward~J Hu, Yelong Shen, Phillip Wallis, Zeyuan Allen-Zhu, Yuanzhi Li, Shean
  Wang, Lu~Wang, and Weizhu Chen.
\newblock Lora: Low-rank adaptation of large language models.
\newblock \emph{arXiv preprint arXiv:2106.09685}, 2021.

\bibitem[Roich et~al.(2022)Roich, Mokady, Bermano, and
  Cohen-Or]{roich2022pivotal}
Daniel Roich, Ron Mokady, Amit~H Bermano, and Daniel Cohen-Or.
\newblock Pivotal tuning for latent-based editing of real images.
\newblock \emph{ACM Transactions on Graphics (TOG)}, 42\penalty0 (1):\penalty0
  1--13, 2022.

\bibitem[Gal et~al.(2022)Gal, Alaluf, Atzmon, Patashnik, Bermano, Chechik, and
  Cohen-Or]{gal2022textual}
Rinon Gal, Yuval Alaluf, Yuval Atzmon, Or~Patashnik, Amit~H. Bermano, Gal
  Chechik, and Daniel Cohen-Or.
\newblock An image is worth one word: Personalizing text-to-image generation
  using textual inversion, 2022.
\newblock URL \url{https://arxiv.org/abs/2208.01618}.

\bibitem[Gimenez et~al.(2016)Gimenez, Robert, Suard, and
  Zreik]{gimenez2016automatic}
Lucile Gimenez, Sylvain Robert, Fr{\'e}d{\'e}ric Suard, and Khaldoun Zreik.
\newblock Automatic reconstruction of {3D} building models from scanned {2D}
  floor plans.
\newblock \emph{Automation in Construction}, 63:\penalty0 48--56, 2016.

\bibitem[Zeng et~al.(2019)Zeng, Li, Yu, and Fu]{zeng2019deep}
Zhiliang Zeng, Xianzhi Li, Ying~Kin Yu, and Chi-Wing Fu.
\newblock Deep floor plan recognition using a multi-task network with
  room-boundary-guided attention.
\newblock In \emph{IEEE ICCV}, pages 9096--9104, 2019.

\bibitem[Pizarro et~al.(2022)Pizarro, Hitschfeld, Sipiran, and
  Saavedra]{pizarro2022automatic}
Pablo~N Pizarro, Nancy Hitschfeld, Ivan Sipiran, and Jose~M Saavedra.
\newblock Automatic floor plan analysis and recognition.
\newblock \emph{Automation in Construction}, 140:\penalty0 104348, 2022.

\bibitem[Wang et~al.(2021)Wang, Li, Zhang, and Shan]{wang2021gfpgan}
Xintao Wang, Yu~Li, Honglun Zhang, and Ying Shan.
\newblock Towards real-world blind face restoration with generative facial
  prior.
\newblock In \emph{{CVPR}}, 2021.

\bibitem[Yang et~al.(2023)Yang, Wang, Gan, Li, Lin, Wu, Duan, Liu, Liu, Zeng,
  and Wang]{Yang_2023_CVPR}
Zhengyuan Yang, Jianfeng Wang, Zhe Gan, Linjie Li, Kevin Lin, Chenfei Wu, Nan
  Duan, Zicheng Liu, Ce~Liu, Michael Zeng, and Lijuan Wang.
\newblock {ReCo}: Region-controlled text-to-image generation.
\newblock In \emph{{CVPR}}, pages 14246--14255, June 2023.

\bibitem[Gan(2022)]{gan2022bim}
Vincent~JL Gan.
\newblock {BIM}-based graph data model for automatic generative design of
  modular buildings.
\newblock \emph{Automation in Construction}, 134:\penalty0 104062, 2022.

\bibitem[Li et~al.(2023)Li, Liu, Wong, Gan, and Cheng]{li2023automated}
Mingkai Li, Yuhan Liu, Billy~CL Wong, Vincent~JL Gan, and Jack~CP Cheng.
\newblock Automated structural design optimization of steel reinforcement using
  graph neural network and exploratory genetic algorithms.
\newblock \emph{Automation in Construction}, 146:\penalty0 104677, 2023.

\bibitem[Ba et~al.(2022)Ba, Lynch, Ploennigs, Schaper, Lohse, and
  Lorenzi]{ba2022automated}
Amadou Ba, Karol Lynch, Joern Ploennigs, Ben Schaper, Christopher Lohse, and
  Fabio Lorenzi.
\newblock Automated configuration of heterogeneous graph neural networks with a
  semantic math parser for {IoT} systems.
\newblock \emph{IEEE IoT J.}, 10\penalty0 (2):\penalty0 1042--1052, 2022.

\bibitem[Chamberlain et~al.(2021)Chamberlain, Rowbottom, Gorinova, Bronstein,
  Webb, and Rossi]{chamberlain2021grand}
Ben Chamberlain, James Rowbottom, Maria~I Gorinova, Michael Bronstein, Stefan
  Webb, and Emanuele Rossi.
\newblock Grand: Graph neural diffusion.
\newblock In \emph{ICML}, pages 1407--1418, 2021.

\bibitem[Shabani et~al.(2023)Shabani, Hosseini, and
  Furukawa]{shabani2023housediffusion}
Mohammad~Amin Shabani, Sepidehsadat Hosseini, and Yasutaka Furukawa.
\newblock Housediffusion: Vector floorplan generation via a diffusion model
  with discrete and continuous denoising.
\newblock In \emph{IEEE/CVF Conf. on Computer Vision and Pattern Recognition},
  pages 5466--5475, 2023.

\bibitem[Gueze et~al.(2023)Gueze, Ospici, Rohmer, and Cani]{gueze2023floor}
Arnaud Gueze, Matthieu Ospici, Damien Rohmer, and Marie-Paule Cani.
\newblock Floor plan reconstruction from sparse views: Combining graph neural
  network with constrained diffusion.
\newblock In \emph{IEEE/CVF Int. Conf. on Computer Vision}, pages 1583--1592,
  2023.

\end{thebibliography}
\end{document}